\documentclass[conference,a4paper]{IEEEtran}

\usepackage[utf8]{inputenc} 
\usepackage[T1]{fontenc}
\usepackage{url}              %
\usepackage{cite}             %

\usepackage[cmex10]{amsmath}  %
\usepackage{mleftright}       %
\mleftright                   %

\usepackage{graphicx}         %
\usepackage{booktabs}         %
\usepackage[mathcal]{euscript}
\usepackage{color}
\usepackage{amssymb,amsfonts,amscd,dsfont,mathrsfs,bbm}
\newcommand{\QED}{\IEEEQED}

\newif\ifcready %
\creadytrue %
\creadyfalse %

\ifcready
\else
\fi

\allowdisplaybreaks

\DeclareMathAlphabet{\mathpzc}{OT1}{pzc}{m}{it} % load "Zapf Chancery" as a math alphabet

\newcommand{\ellh}{\ell_\hinge}
\newcommand{\hinge}{\mathrm{hinge}}

\newcommand{\T}{\mathrm{T}}
\newcommand{\defeq}{\triangleq}
\newcommand{\ev}{{\mathbb{E}}}

\newcommand{\E}[1]{\ev\left[#1\right]}
\newcommand{\bE}[1]{\ev\bigl[#1\bigr]}

\newcommand{\Ed}[2]{\ev_{#1}\left[#2\right]}

\newcommand{\bbabs}[1]{\left|#1\right|}
\newcommand{\norm}[1]{\|#1\|}
\newcommand{\bnorm}[1]{\bigl\|#1\bigr\|}
\newcommand{\bbnorm}[1]{\left\|#1\right\|}

\newcommand{\ip}[2]{\langle#1,#2\rangle}
\newcommand{\bip}[2]{\bigl\langle#1,#2\bigr\rangle}
\newcommand{\fip}[2]{\left\langle#1,#2\right\rangle}

\newcommand{\eps}{\epsilon}

\newcommand{\X}{\mathcal{X}}
\newcommand{\cF}{\mathcal{F}}
\newcommand{\cG}{\mathcal{G}}

\newcommand{\cI}{\mathcal{I}}
\newcommand{\cX}{\X}
\newcommand{\Z}{\mathcal{Z}}
\newcommand{\cV}{\mathcal{V}}
\newcommand{\cZ}{\Z}

\newcommand{\Pt}{\tilde{P}}

\newcommand{\Y}{\mathcal{Y}}
\newcommand{\cY}{\Y}

\newcommand{\kb}{\bar{k}}

\newcommand{\ft}{\tilde{f}}
\newcommand{\wt}{\tilde{w}}
\newcommand{\nut}{\tilde{\nu}}
\newcommand{\thetat}{\tilde{\theta}}

\newcommand{\kr}{\mathpzc{k}}
\newcommand{\krf}[1]{\kr_{{\;}#1}} % feature kernel
\newcommand{\krs}{\kr^*} % maximal correlation kernel
\newcommand{\krt}{\tilde{\kr}} % zero-mean
\newcommand{\lra}{\leftrightarrow} % kernel mat

\DeclareMathOperator{\spn}{span}

\DeclareMathAlphabet{\mathbbb}{U}{bbold}{m}{n}  % bb bold numbers
\DeclareMathOperator{\sign}{sgn}
\DeclareMathOperator{\diag}{diag}
\newcommand{\kron}{\mathbbb{1}}

\newtheorem{definition}{Definition}%[section]
\newtheorem{theorem}{Theorem}%[section]
\newtheorem{property}{Property}%[section]
\newtheorem{proposition}{Proposition}%[section]
\newtheorem{fact}{Fact}
\newtheorem{corollary}{Corollary}%[thm]
 \newtheorem{remark}{Remark}%[section]

\newcommand{\ds}{\displaystyle}

\newcommand{\appref}[1]{Appendix~\ref{#1}}

\newcommand{\thmref}[1]{Theorem~\ref{#1}}
\newcommand{\propref}[1]{Proposition~\ref{#1}}
\newcommand{\proptyref}[1]{Property~\ref{#1}}
\newcommand{\corolref}[1]{Corollary~\ref{#1}}
\newcommand{\factref}[1]{Fact~\ref{#1}}
\newcommand{\defref}[1]{Definition~\ref{#1}}

\newcommand{\Hs}{\mathscr{H}} % Hscore

\newcommand{\yh}{{\hat{y}}}
\newcommand{\fh}{{\hat{f}}}
\newcommand{\gh}{{\hat{g}}}

\newcommand{\MAP}{{\mathsf{MAP}}}
\newcommand{\LR}{{\mathsf{LR}}}
\newcommand{\SVM}{{\mathsf{SVM}}}
\newcommand{\wsvm}{w_{\SVM}}
\newcommand{\bsvm}{b_{\SVM}}
\newcommand{\wlr}{w_{\LR}}
\newcommand{\blr}{b_{\LR}}

\newcommand{\yhkr}[1]{\yh^{(#1)}}
\newcommand{\pkr}[1]{P^{(#1)}}
\newcommand{\lsvm}{L_{\SVM}}
\newcommand{\lsvms}{\hat{L}} % surrogated svm loss 

\newcommand{\rhomax}{\varrho}
\DeclareMathOperator*{\argmax}{arg\,max}
\DeclareMathOperator*{\argmin}{arg\,min}

\DeclareMathOperator{\sigmoid}{sigmoid}

\newcommand{\La}{\Lambda}
\newcommand{\funcs}{\cF}
\newcommand{\spc}[1]{\funcs_{#1}}
\newcommand{\proj}[2]{\Pi\left(#1;{#2}\right)}   % \proj{f}{F} proj f onto F

\begin{document}

\title{Kernel Subspace and Feature Extraction} 

\author{%
  \IEEEauthorblockN{Xiangxiang Xu and Lizhong Zheng}
  \IEEEauthorblockA{
    Dept. EECS, MIT\\
    Cambridge, MA 02139, USA\\
    Email: \{xuxx, lizhong\}@mit.edu}
}%

\maketitle

\begin{abstract}
  We study kernel methods in machine learning from the perspective of feature subspace. We establish a one-to-one correspondence between feature subspaces and kernels and propose an information-theoretic measure for kernels. In particular, we construct a kernel from Hirschfeld--Gebelein--R\'{e}nyi maximal correlation functions, coined the maximal correlation kernel, and demonstrate its information-theoretic optimality. We use the support vector machine (SVM) as an example to illustrate a connection between kernel methods and feature extraction approaches. We show that the kernel SVM on maximal correlation kernel achieves minimum prediction error. Finally, we interpret the Fisher kernel as a special maximal correlation kernel and establish its optimality.
\end{abstract}

\section{Introduction}
One main objective of machine learning is to obtain useful information from often high-dimensional data. To this end, it is a common practice to extract meaningful feature representations from original data and then process features \cite{storcheus2015survey}. Neural networks \cite{lecun1998gradient} and kernel methods \cite{mika1999fisher, bach2002kernel, hofmann2008kernel, scholkopf2000kernel} are two of the most representative approaches to map data into feature space. In neural networks, the features are represented as the outputs of hidden neurons in the network. In contrast, the feature mapping in kernel methods is defined by the used kernel, which is used implicitly and is often infinite dimensional. While kernel approaches require much fewer parameters and can obtain good empirical performance on certain tasks \cite{cortes1995support}, the performance significantly relies on the choice of kernels. With many attempts to investigate kernel methods
 \cite{platt1999probabilistic, xu2010reproducing,
 scholkopf2000kernel}, there still lacks a theoretical understanding of the mechanism behind kernel methods, which restricts their applications on complicated data.

On the other hand, the  feature extraction in  deep neural networks has been studied recently by information-theoretic and statistical analyses \cite{xu2022information, huang2019universal}. For example, it was shown in \cite{xu2022information} that, the feature extracted by deep neural networks coincides with the most informative feature, which is essentially related to the classical Hirschfeld--Gebelein--R\'{e}nyi (HGR) maximal correlation problem \cite{hirschfeld1935connection, gebelein1941statistische, renyi1959measures}.
Such theoretical characterizations provide a better understanding of existing algorithms and have been shown useful in designing algorithms for multimodal learning tasks \cite{xu2022multivariate}.

In this paper, our goal is to characterize kernel methods from the perspective of feature subspace and reveal its connection with other learning approaches. We first introduce the associated kernel with each given feature subspace, which we coin the \emph{projection kernel}, to  establish a correspondence between kernel operations and geometric operations in feature subspaces. This connection allows us to study kernels methods via analyzing the corresponding feature subspaces. Specifically, we propose an information-theoretic measure for projection kernels, and demonstrate that the information-theoretically optimal kernel can be constructed from the HGR maximal correlation functions, coined the \emph{maximal correlation kernel}. We further demonstrate that the support vector machine (SVM) with maximal correlation kernel can obtain the minimum prediction error, which justifies its optimality in learning tasks. Our analysis also reveals connections between SVM and other classification approaches including neural networks. Finally, we interpret the Fisher kernel, a classical kernel induced from parameterized distribution families \cite{jaakkola1999exploiting}, as a special case of maximal correlation kernels, thus demonstrating its optimality.

\section{%
  Preliminaries and Notations}
\label{sec:pf}

Throughout this paper, we use $X, Y$ to denote two random variables with alphabets $\cX, \cY$, and denote their joint distribution and marginals as $P_{X, Y}$ and $P_X, P_Y$, respectively. We also use $\E{\cdot}$ to denote the expectation with respect to $P_{X, Y}$.

\subsection{Feature Space} %
We adopt the notation convention introduced in \cite{xu2022multivariate}, and %
let $\spc{\cX}\defeq \{\cX \to \mathbb{R}\}$ denote the feature space formed by the (one-dimensional) features of $X$, with the geometry defined as follows. The inner product $\ip{\cdot}{\cdot}_{\spc{\cX}}$ on $\spc{\X}$ is defined as  $\ip{f_1}{f_2}_{\spc{\cX}} \defeq \Ed{P_X}{f_1(X)f_2(X)}$ for $f_1, f_2 \in \spc{\cX}$. This induces a norm $\norm{\,\cdot\,}_{\spc{\cX}}$ with $ \norm{f}_{\spc{\cX}} \defeq   \sqrt{\ip{f}{f}_{\spc{\cX}}}$ for $f \in \spc{\cX}$. Then, for given $f \in \spc{\cX}$ and subspace $\cG$ of $\spc{\cX}$, we denote the projection of $f$ onto $\cG$ as
\begin{align}
  \proj{f}{\cG} \defeq  \argmin_{h \in \cG} \norm{h - f}_{\spc{\cX}}.
  \label{eq:def:proj}
\end{align}

In addition, for a $d$-dimensional feature $f =  (f_1, \dots, f_d)^\T \colon \cX \to \mathbb{R}^d$, we use $\spn\{f\} \defeq \spn\{f_1, \dots, f_d\}$ to denote the subspace spanned by all dimensions. We also use $\ft$ to denote the centered $f$, i.e., $\ft(x) \defeq f(x) - \Ed{P_X}{f(X)}$, and denote $\Lambda_f \defeq \Ed{P_X}{f(X)f^\T(X)}$.
\subsection{Kernel}
Given $\cX$, $\kr \colon \cX \times \cX \to \mathbb{R}$ is a kernel on $\cX$, if for all finite subset $\cI \subset \cX$, the $|\cI|$ by $|\cI|$ matrix $[\kr(x, x')]_{x \in \cI, x' \in \cI}$ is positive semidefinite.
For each kernel $\kr$, we define the associated functional operator $\tau \colon \spc{\cX} \to \spc{\cX}$ as
\begin{align}
  [\tau(f)](x) \defeq \Ed{P_X}{\kr(X, x) f(X)},
  \label{eq:def:tau}
\end{align}
and we use  $\kr \lra \tau$ to denote the correspondence between $\kr$ and $\tau$. Furthermore, we define the centered kernel $\krt \colon \cX \times \cX \to \mathbb{R}$ as
\begin{align}
  \krt(x, x') \defeq \kr(x, x') - \kb(x) - \kb(x') + 
\Ed{P_X}{\kb(X)},
\label{eq:def:krt}
\end{align}
where we have defined $\kb\colon (x \mapsto \Ed{P_X}{\kr(X, x)}) \in \spc{\cX}$.

The following fact is the basis of the kernel trick in learning algorithms.
\begin{fact}
  \label{fact:kernel:mapping}
  For each given kernel $\kr$, there exist an inner product space $\cV$ with the inner product $\ip{\cdot}{\cdot}_{\cV}$, and a mapping $\nu\colon \cX \to \cV$, such that $\kr(x, x')
  = \ip{\nu(x)}{\nu(x')}_{\cV}$.%
\end{fact}

\begin{remark}
Suppose $\nu$ is one mapping for $\kr$ satisfying \factref{fact:kernel:mapping}. Then for the centered kernel $\krt$ [cf. \eqref{eq:def:krt}], %
 we have $\krt(x, x') = \ip{\nut(x)}{\nut(x')}_{\cV}$, where $\nut(x) \defeq \nu(x)- \Ed{P_X}{\nu(X)}$.
\end{remark}

In addition, we introduce the kernelized discriminative model (KDM) as follows.
\begin{definition}[Kernelized Discriminative Model]
\label{def:kdm}
For each kernel $\kr$, we define its associated kernelized discriminative model $P^{(\kr)}_{Y|X}$ as
  \begin{align}
    \pkr{\kr}_{Y|X}(y|x) \defeq P_Y(y) \left( 1 + \E{\krt(X, x)\middle|Y = y}\right).
    \label{eq:pkr:def}
  \end{align}
  Then, we use $\yhkr{\kr}$ to denote the  maximum a posteriori (MAP) estimation induced from KDM $\pkr{\kr}_{Y|X}$, i.e.,
\begin{align}
  \yhkr{\kr}(x) &\defeq \argmax_{y\in \cY} \pkr{\kr}_{Y|X}(y|x).
  \label{eq:kr:pred}
\end{align}
\end{definition}
The KDM can be regarded as a generalized probability distribution, since
we have $\sum_{y \in \cY} P^{(\kr)}_{Y|X}(y|x) = 1$ for all $x \in \cX$ while $\pkr{\kr}_{Y|X}(y|x)$ can sometimes take negative values.

\subsection{
 Modal Decomposition,
  Maximal Correlation, and H-score}
We first introduce the modal decomposition of joint distribution $P_{X, Y}$ \cite{huang2019universal, xu2022multivariate}.
\begin{proposition}[Modal Decomposition \cite{huang2019universal}]
\label{prop:modal}
  For given $P_{X, Y}$, there exists $K \leq \min\{|\cX|, |\cY|\} - 1$, such that
\begin{align}
  P_{X, Y}(x, y) = P_X(x)P_Y(y) \left(1 + \sum_{i = 1}^K \sigma_i f^*_i(x) g^*_i(y) \right),
  \label{eq:modal}
\end{align}
where $\sigma_1 \geq \sigma_2 \geq \sigma_K > 0$, and %
$\E{f^*_i(X)f^*_j(X)} = \E{g^*_i(Y)g^*_j(Y)} = \kron_{\{i = j\}}$ for all $1 \leq i, j \leq K$ , where $\kron_{\{\cdot\}}$ denotes the indicator function. 
\end{proposition}

It can be shown that $(f_i^*, g_i^*)$ pairs are the most correlated function pairs of $X$ and $Y$, referred to as maximal correlation functions.
We also denote $\rhomax \defeq \sigma_1$, known as the HGR maximal correlation \cite{hirschfeld1935connection, gebelein1941statistische, renyi1959measures} of $X$ and $Y$, and define the $K$-dimensional feature  $f^*(x) \defeq [f^*_1(x), \dots, f_K^*(x)]^\T$. In particular, when $Y$ is binary, we have $f^* = f_1^* \in \spc{\cX}$. %

It has been shown in \cite{huang2019universal}
 that the maximal correlation functions $f_i^*, i = 1, \dots, K$ are the optimal features of $X$ in inferring or estimating $Y$. In general, given a $d$-dimensional feature $f$ of $X$, the effectiveness of $f$ in inferring or estimating $Y$ can be measured by its \emph{H-score} \cite{xu2022information, huang2019universal}, defined as
\begin{align}
  \Hs(f) \defeq \frac{1}{2}\cdot \E{\bbnorm{\E{\Lambda_{f}^{-\frac12}\ft(X)\middle|Y}}^2},
  \label{eq:def:h}
\end{align}
where $\ft(x) \defeq f(x) - \E{f(X)}$. It can be verified that for all $d$ and $f \colon \cX \to \mathbb{R}^d$, we have
\begin{align}
  \Hs(f) \leq \Hs(f^*) = \frac12\sum_{i = 1}^K \sigma_i^2,
  \label{eq:h:max}
\end{align}
where $\sigma_1, \dots, \sigma_K$ are as defined in \eqref{eq:modal}.

\subsection{Binary Classification}
We consider the binary classification problem which predicts binary label $Y$ from the data variable $X$. For convenience, we assume $Y$ takes values from $ \cY \defeq \{-1, 1\}$.

Suppose the training dataset contains $n$ sample pairs $\{(x_i, y_i)\}_{i = 1}^n$ of $(X, Y)$, 
and let $P_{X, Y}$ denote the corresponding empirical distribution, i.e.,
\begin{align}
  P_{X, Y}(x, y)\defeq\frac1n\sum_{i = 1}^n\kron_{\{x_i = x, y_i = y\}}.
  \label{eq:emp:dist}
\end{align}

\subsubsection{Support Vector Machine}

The support vector machine (SVM) solves binary classification tasks by finding the optimal hyperplane that separates two classes with maximum margin \cite{cortes1995support}. Given $d$-dimensional feature mapping  $f \colon \cX \to \mathbb{R}^{d}$, the loss for SVM based on $f$ can be written as
\begin{align}
  & \lsvm(f, w, b; \lambda)
    \notag\\
 &\quad\defeq \Ed{P_{X, Y}}{ \ellh(Y, \ip{w}{f(X)} + b)} + \frac{\lambda}2\cdot \|w\|^2,
\end{align}
where $w, b \in \mathbb{R}^d$ are the parameters of the hyperplane, where $\lambda > 0$ is a hyperparameter of SVM, and where $\ellh \colon \Y \times \mathbb{R} \to \mathbb{R}$ denotes the hinge loss, defined as $ \ellh(y, z) \defeq (1 - yz)^+ $ with $x^+ \defeq \max\{0, x\}$. 

Moreover, let $\ds (\wsvm, \bsvm) \defeq \argmin_{w, b} \lsvm(f, w, b; \lambda)$
and  $ \lsvm^*(f; \lambda) \defeq \lsvm(f, \wsvm, \bsvm; \lambda)$ denote the optimal parameters and the value of loss function, respectively. Then, the prediction of SVM is
\begin{align}
  \yh_{\SVM}(x; f, \lambda) \defeq \sign(\ip{\wsvm}{f(x)} + \bsvm),
  \label{eq:yh:svm}
\end{align}
where $\sign(\cdot)$ denotes the sign function.

Specifically, for a given kernel $\kr$, the prediction of the corresponding \emph{kernel SVM} is\footnote{It is worth mentioning that
  the practical implementation of kernel SVM is typically done by solving a dual optimization problem without explicitly using $\nu$. %
  See \cite[Section 12]{hastie2009elements} for detailed discussions.}
 $ \yhkr{\kr}_{\SVM}(x; \lambda) \defeq \yh_{\SVM}(x; \nu, \lambda),$
where $\nu$ is any mapping given by \factref{fact:kernel:mapping}.

\subsubsection{Logistic Regression and Neural Networks}
Given $d$-dimensional feature $f$ of $X$, the discriminative model of logistic regression is %
 $ \Pt_{Y|X}(y|x; f, w, b) \defeq \sigmoid (y\cdot (\ip{w}{ f(x)} + b))$,
where $w \in \mathbb{R}^d, b \in \mathbb{R}$ are the weight and bias, respectively, and where $\sigmoid(\cdot)$ is defined as $\sigmoid(x) \defeq \frac1{1+ \exp(-x)}$. 

Then, the loss of logistic regression is
$L_{\LR}(f, w, b) \defeq -\E{\log \Pt_{Y|X}(Y|X; f, w, b)}$, and the optimal parameters  $\wlr, \blr$ are learned by minimizing the loss, i.e., $\ds (\wlr, \blr) \defeq \argmin_{w, b} L_{\LR}(f, w, b)$. The resulting decision rule is
\begin{align}
  \yh_{\LR}(x; f) 
  &\defeq \argmax_{y\in \cY} \Pt_{Y|X}(y|x; f, \wlr, \blr) \notag\\
  &= \sign\left(\ip{\wlr}{ f(x)} + \blr\right).
  \label{eq:yh:lr}
\end{align}

The logistic regression is often used as the classification layer for multi-layer neural networks, where $w$ and $b$ correspond to weights and the bias term, respectively. In this case, the feature mapping $f(\cdot)$ also takes a parameterized form, and the parameters of $f(\cdot)$ are jointly learned with $w$ and $b$.

\ifcready
Due to the space limitations, we omit most proofs %
in the rest of this paper, but refer the readers to the extended version of this paper \cite{xu2023kernel} for the details.
\fi

\section{Projection Kernel and Informative Features}
In this section, we introduce a one-to-one correspondence between kernels and feature subspaces, and then characterize the informativeness of kernels by investigating the features in the associated subspaces.
\subsection{Projection Kernel and Feature Subspace}
We first introduce a family of kernels with one-to-one correspondence to feature subspace.

\begin{definition}[Projection Kernel]
  Let $\cG$ denote a $d$-dimensional subspace of $\spc{\cX}$ with a basis $\{f_1, \dots, f_d\}$. We use $\krf{\cG}\colon \cX \times \cX \to \mathbb{R}$ to denote the projection kernel associated with $\cG$, defined as  
   $ \krf{\cG}(x, x') \defeq f^\T(x) \Lambda_f^{-1} f(x')$,
  where we have defined $f \defeq (f_1, \dots, f_d)^{\T}$ and $\Lambda_f \defeq \E{f(X)f^\T(X)}$.
\end{definition}

With slight abuse of notation, we also dnenote $\krf{f} \defeq \krf{\spn\{f\}}$, the projection kernel associated with $\spn\{f\}$.

Note that $\krf{\cG}$ is a valid kernel function, and the corresponding $\nu$ mapping in \factref{fact:kernel:mapping} can be chosen as $\nu(x) = [f_1(x), \dots, f_d(x)]^\T$ for any orthonormal basis $\{f_1, \dots, f_d\}$  of $\cG$.  It turns out that the functional operators associated with projection kernels are projection operators in the feature space, which we formalize as follows. 
\ifcready\else
 A proof is provided in \appref{app:propty:proj}.
\fi

\begin{property}
\label{propty:proj}
Let $\tau \lra \krf{\cG}$ denote the operator corresponding to subspace $\cG$ [cf. \eqref{eq:def:tau}], then we have  $\tau(f) = \proj{f}{\cG}$ for all $f \in \spc{\cX}$.
\end{property}

Therefore, given a projection kernel $\kr$, the associated subspace can be represented as $\{f \in \spc{\cX} \colon \tau(f) = f\}$, where  $\tau \lra  \kr$ is the associated operator. This establishes a one-to-one correspondence between projection kernels and feature subspaces. 

\subsection{H-score and Informative Features}
The projection kernel provides a connection between feature subspace and kernel, from which we can characterize subspace $\cG$ in terms of the corresponding kernel $\krf{\cG}$. Specifically, we can represent the H-score  [cf. \eqref{eq:def:h}] of a feature $f$ in terms of the projection kernel $\krf{f}$, formalized as follows. 
\ifcready\else
 A proof is provided in \appref{app:prop:h:subspace}.
\fi

\begin{proposition}
  \label{prop:h:subspace}
For all $f$ with $\spn\{f\} = \cG$, we have $\Hs(f) = \frac12 \cdot \left(\Ed{P_{X X'}}{\krf{\cG}(X, X')} - \Ed{P_{X}P_{X'}}{\krf{\cG}(X, X')}\right)$, where we have defined $X'$ such that the joint distribution of $X$ and $X'$ is
\begin{align}
  P_{X X'}(x, x') \defeq \sum_{y \in \Y} P_{Y}(y)P_{X|Y=y}(x)P_{X|Y=y}(x').
  \label{eq:p:xx}
\end{align}
\end{proposition}

With slight abuse of notation, we can use $\Hs(\cG)$ to denote the H-score corresponding to feature subspace $\cG$. In particular, we have the following characterization of $\Hs(\cG)$ when  $Y$ is binary. 
\ifcready\else
 A proof is provided in \appref{app:prop:info:f}.
\fi

\begin{proposition}
  \label{prop:info:f}
  Suppose $Y$ is binary, and $f^*$ is the maximal correlation function of $P_{X, Y}$. Then, for each subspace $\cG$ of $\spc{\cX}$, we have
  \begin{align}
    \Hs(\cG) &= \frac{\rhomax^2}2 \cdot \bnorm{\proj{f^*}{\cG}}_{\spc{\cX}}^2%
                 = \max_{f\in \cG} \Hs(f)
                 = \Hs(\proj{f^*}{\cG}).
    \label{eq:f:cG}
  \end{align}
\end{proposition}

From \propref{prop:info:f}, $\Hs(\cG)$ depends only on the projection of $f^*$ onto $\cG$, which is also the most informative feature in $\cG$. In addition, note that since $\norm{f^*}_{\spc{\cX}} = 1$,  $\bnorm{\proj{f^*}{\cG}}_{\spc{\cX}}^2$ is also the cosine value of the principal angle between $f^*$ and $\cG$. Therefore, we can interpret the H-score as a measure of the principal angle between the optimal feature $f^*$ and the given subspace.

\subsection{Maximal Correlation Kernel}
Note that from \eqref{eq:h:max}, $\Hs(f)$ is maximized when $f$ takes the maximal correlation function $f^*$. Therefore, the subspace $\spn\{f^*\}$ (and thus projection kernel $\krf{f^*}$) is optimal in terms of the  H-score measure. We will denote $\krs \defeq \krf{f^*}$, referred to as the \emph{maximal correlation kernel}. 

Specifically, the KDM (cf. \defref{def:kdm}) of maximal correlation kernel $\krs$ coincides with the underlying conditional distribution $P_{Y|X}$, demonstrated as follows. 
\ifcready\else
 A proof is provided in \appref{app:propty:kdm}.
\fi

\begin{property} 
\label{propty:kdm}
For all $x$ and $y$, we have $P_{Y|X}(y|x) = P^{(\krs)}_{Y|X}(y|x)$ and  $\yh^{(\krs)}(x) = \yh_{\MAP}(x)$, where
$\yh_\MAP$ denotes the MAP estimation, i.e.,
 \begin{align}
 \yh_{\MAP}(x) \defeq \argmax_{y \in \cY} P_{Y|X}(y|x).
  \label{eq:map}
\end{align}
\end{property}

As we will develop in the next section, 
the maximal correlation kernel also achieves the optimal performance in support vector machine. %

 \section{Support Vector Machine Analysis%
 }
\label{sec:prac}

In this section, we investigate support vector machine, a representative kernel approach for binary classification. Let $(X, Y)$ denote the training data and corresponding label taken from  $\cY = \{-1, 1\}$, with $P_{X, Y}$ denoting the empirical distribution as defined in \eqref{eq:emp:dist}. Throughout this section, we will focus on the balanced dataset with
\begin{align}
  P_Y(-1) = P_Y(1) = \frac12.
  \label{eq:balanced}
\end{align}
It can be verified that in this case, the MAP estimation [cf. \eqref{eq:map}] can be expressed in terms of maximal correlation function.
\ifcready\else
 A proof is provided in \appref{app:propty:map}.
\fi
\begin{property}
  \label{propty:map}
  Under assumption \eqref{eq:balanced}, we can express the MAP estimation as $\yh_{\MAP}(x) = \sign(f^*(x)) $ for all $x \in \cX$, where $f^* \in \spc{\cX}$ is the maximal correlation function of $P_{X, Y}$.
\end{property}

\subsection{SVM on Given Features}
We first consider the SVM algorithm applied on a given feature representation $f(X) \in \mathbb{R}^d$, which can also be regarded as the kernel SVM on kernel $\kr(x, x') = \ip{f(x)}{f(x')}$. 

To begin, for each given feature $f$ and $\lambda > 0$, let us define 
\begin{align*}
  \lsvms(f; \lambda) \defeq 1 - \frac{1}{2\lambda} \cdot  \bbnorm{\E{f(X)Y}}^2.
\end{align*}
\ifcready
Then we have the following characterization.
\else
Then we have the following characterization, a proof of which is provided in \appref{app:thm:svm}.
\fi

\begin{theorem}
  \label{thm:svm}    
  For all given feature $f$ and $\lambda \geq 0$, we have
  \begin{align}
    \lsvms(f;\lambda) \leq \lsvm^*(f;\lambda) \leq \lsvms(f;\lambda) + \left(\frac{\lambda_\T}{\lambda} - 1\right)^{+},
  \end{align}
  where we have defined
  $\lambda_{\T} \defeq M \cdot \bbnorm{\E{f(X)Y}}$ and $M \defeq \max_{x \in \cX} \bnorm{\ft(x)}$, with $\ft(x) \defeq f(x) - \E{f(X)}$, and where $x^+ \defeq \max\{0, x\}$.

  Specifically, when $\lambda \geq \lambda_{\T}$, we have
  $\lsvm^*(f;\lambda) = \lsvms(f;\lambda)$, which can be achieved by
  \begin{align}
    \wsvm = \frac{1}{\lambda} \cdot \E{f(X) Y}, \quad \bsvm = - \ip{\wsvm}{\E{f(X)}},
  \end{align}
  and the resulting SVM prediction is 
  \begin{align}
    \yh_{\SVM}(x; f, \lambda) &= \sign\left(\fip{\bE{\ft(X) Y}}{\ft(x)}\right)\label{eq:svm:decision}\\
    &= \argmin_{y \in \cY} \norm{f(x) - \E{f(X)|Y = y}}.\label{eq:svm:decision:ncc}
  \end{align}
\end{theorem}

From \thmref{thm:svm}, when $\lambda \geq \lambda_\T$, the SVM decision $\yh_{\SVM}(x; f, \lambda)$  does not depend on the value of $\lambda$. In the remaining, we will focus on the regime where $\lambda \geq \lambda_\T$, and drop the $\lambda$ in expressions whenever possible, e.g., we simply denote $\yh_{\SVM}(x; f, \lambda)$ by $\yh_{\SVM}(x; f)$. As we will see soon, SVM can still obtain minimum prediction error in this regime, by using a good feature mapping $f$ (or equivalently, a good kernel).
From \eqref{eq:svm:decision:ncc}, the SVM prediction can be interpreted as a nearest centroid classifier, where decision is based on comparing the distance between $f(x)$ and the class centroids $\E{f(X)|Y = y}$, $y \in  \cY$. 
In addition, from
  \begin{align*}
    \E{f(X)Y} 
    &= \E{Y \cdot \E{f(X)|Y}}\\
    &= \frac12 \left(\E{f(X)|Y=1} - \E{f(X)|Y=-1}\right),
  \end{align*}
  we can interpret the SVM loss $\lsvm^*= \lsvms$ as measuring the distance between two class centroids.

Furthermore, when $f$ is one-dimensional feature, we can rewrite \eqref{eq:svm:decision} as
\begin{align*}
  \yh_{\SVM}(x; f) = \sign\left(\fip{\bE{\ft(X) Y}}{\ft(x)}\right) = \sign\left(\fh(x)\right),
\end{align*}
where $\fh \defeq \proj{f^*}{\spn\{\ft\}}$. Therefore, the decision rule depends only the projection of $f^*$ onto the subspace $\spn\{\fh\}$, which is also the most informative features on the subspace (cf. \propref{prop:info:f}). Later on we will see a similar geometric illustration of kernel SVM.

Moreover, we can establish a connection between SVM loss and the H-score measure, formalized as the following corollary. 
\ifcready
\else
A proof is provided in \appref{app:cor:svm}.
\fi
  \begin{corollary}
    \label{cor:svm}
    Suppose $\lambda \geq \lambda_\T$, then we have
    \begin{align*}
     1 - \frac{r_{\max}}{\lambda} \cdot \Hs(\ft) \leq \lsvm^*(f;\lambda)\leq 1 - \frac{r_{\min}}{\lambda} \cdot \Hs(\ft),
    \end{align*}
    where  $r_{\max}$ and $r_{\min}$ denote the maximum and minimum positive eigenvalues of the covariance matrix $\Lambda_{\ft}$, respectively. Specifically, if $\Lambda_{\ft} = I$, then we have
      $\lsvm^*(f;\lambda)= 1 - \lambda^{-1} \cdot \Hs(\ft).$
  \end{corollary}

As a result, for each normalized feature $f$ with covariance matrix $\Lambda_{\ft} = I$, the SVM loss $\lsvm^*$ measures the informativeness of $f$ in inferring the label $Y$.

\subsection{Kernel SVM}
In practice, instead of applying SVM on a given or manually designed feature $f$, it is more often to directly implement SVM on a kernel $\kr$. Similar to \thmref{thm:svm}, we have the following characterization, from which we can interpret KDM as a probabilistic output for kernel SVM.

\begin{theorem}
  \label{thm:ksvm}
  For each given kernel $\kr$, there exists a constant $\lambda_\T > 0$, such that when $\lambda \geq \lambda_\T$, the SVM prediction is  $\yhkr{\kr}_{\SVM}(x) = \sign\left([\tau(f^*)](x)\right)$, where $\tau \lra \krt$ is the  operator associated with centered kernel $\krt$ [cf. \eqref{eq:def:tau} and \eqref{eq:def:krt}]. In addition, the SVM prediction coincides with the KDM prediction (cf. \defref{def:kdm}) obtained from $\kr$, i.e., we have  $\yh^{(\kr)}_{\SVM}(x) = \yh^{(\kr)}(x)$ for all $x \in \cX$.%
\end{theorem}
\begin{IEEEproof}
Let $\cV$ and $\nu \colon \cX \to \cV$ denote the inner product space and mapping associated with kernel $\kr$ (cf. \factref{fact:kernel:mapping}), and let $\nut(x) \defeq \nu(x) - \Ed{P_X}{\nu(X)}$. Then, we have
  \begin{align}
     \ip{ \E{\nut(X)Y}}{\nut(x)}_{\cV}
    &=  \E{\ip{ \nut(X)}{\nut(x)}_{\cV}\cdot Y}\notag\\
    &=  \E{\krt(X, x)\cdot Y},
    \label{eq:nut:krt}
  \end{align}
  which can be rewritten as
  \begin{align*}
    &\E{\krt(X, x) \cdot Y}\\
    &\qquad= \Ed{P_{X, Y}}{\krt(X, x) \cdot Y}\\
    &\qquad=     \Ed{P_{X}P_{ Y}}{\krt(X, x) \cdot Y \cdot (1 + \rhomax \cdot f^*(X) \cdot Y)}\\
    &\qquad=     \E{\krt(X, x)}\cdot \E{ Y} + \rhomax \cdot  \E{\krt(X, x) f^*(X)} \cdot \E{Y^2}\\
    &\qquad= \rhomax\cdot \E{\krt(X, x) f^*(X)}\\
    &\qquad=  \rhomax\cdot [\tau(f^*)](x),
  \end{align*}
  where to obtain the second equality we have used the modal decomposition of $P_{X, Y}$\ifcready.\else~(cf. \factref{fact:modal}).\fi

  Hence, from \thmref{thm:svm} we obtain
  \begin{align*}
    \yhkr{\kr}_{\SVM}(x) = \yh_{\SVM}(x; \nu) &= \sign\left(\ip{ \E{\nut(X)Y}}{\nut(x)}\right)\\
                                              &= \sign\left(\E{\krt(X, x)\cdot Y}\right)\\
                                              &=  \sign( [\tau(f^*)](x)).
  \end{align*}
  
  It remains only to establish the equivalence between $\yhkr{\kr}_{\SVM}$ and KDM decision  $\yhkr{\kr}$. To this end, note that from \eqref{eq:pkr:def} and the balanced dataset assumption \eqref{eq:balanced}, we have
  \begin{align*}
    P^{(\kr)}_{Y|X}(y|x) 
    &= P_Y(y) \left( 1 + \E{\krt(X, x)\middle|Y = y}\right)\\
    &= \frac12 \left( 1 + y\cdot\E{\krt(X, x) Y}\right)
  \end{align*}
  for all $x\in \cX, y \in \cY$.

  Hence, for all $x \in \cX$, 
  \begin{align*}
    \yhkr{\kr}(x) = \argmax_{y\in \cY}{P^{(\kr)}_{Y|X}(y|x) } &= \sign\left(\E{\krt(X, x) Y}\right)\\
    &= \yhkr{\kr}_{\SVM}(x),
  \end{align*}
  which completes the proof.
\end{IEEEproof}

From \thmref{thm:ksvm}, the final decision $\yhkr{\kr}_{\SVM}$ depends on $\kr$ only through the centered kernel $\krt$. Moreover, compare \thmref{thm:ksvm} with \proptyref{propty:map}, kernel SVM prediction differs from MAP only in applying the operator $\tau$ on $f^*$. In particular, when the maximal correlation function $f^*$ is an eigenfunction of the corresponding operator $\tau \lra \krt$, i.e., $\tau(f^*) = c \cdot f^*$ for some $c > 0$, the SVM prediction coincides with the MAP prediction, i.e., $\yhkr{\kr}_{\SVM}(x) = \yh_{\MAP}(x)$ for all $x \in \cX$.
If we restrict our attention to  projection kernels, the kernel SVM decision can be further interpreted as  a projection operation on the associated subspace. To see this, let $\cG$ denote a feature subspace of $\spc{\cX}$ spanned by zero-mean features, then from \thmref{thm:svm} and \propref{prop:info:f}, the kernel SVM loss for $\krf{\cG}$ is
 \begin{align*}
 1 - \frac{1}{\lambda} \cdot  \Hs(\cG) = 1 - \frac{\rhomax^2}{2\lambda} \cdot \bnorm{\proj{f^*}{\cG}}^2_{\spc{\cX}}, 
 \end{align*}
which measures the principal angle between $f^*$ and $\cG$. In addition, the decision rule can be expressed as
\begin{align}
  \yhkr{\krf{\cG}}_{\SVM}(x) = \sign([\proj{f^*}{\cG}](x)),%
  \label{eq:yh:ksvm}
\end{align}
From \propref{prop:info:f}, $\proj{f^*}{\cG}$ is also the most informative feature in $\cG$. Therefore, kernel SVM on $\krf{\cG}$ is equivalent to first extracting the most informative feature in $\cG$, and then using the extracted feature to make decision. %

\subsection{Relationship to Other Classification Approaches}
\subsubsection{Maximum a Posteriori (MAP) Estimation}
From \eqref{eq:yh:ksvm}, when the maximal correlation kernel $\krs$ is applied, the kernel SVM decision is $\sign(f^*(x))$, which coincides with the MAP prediction (cf. \proptyref{propty:map}). Since MAP achieves the minimum prediction error, kernel SVM on the maximal correlation kernel also obtains the minimum prediction error.

\subsubsection{Logistic Regression and Neural Networks}
We have interpreted SVM as extracting the most informative feature, where the informativeness is measured by H-score. 
The analysis in \cite{xu2022information} has shown that %
logistic regression is also equivalent to maximizing the H-score, when $X$ and $Y$ are weakly independent.  Indeed, we can show that SVM and logistic regression lead to the same prediction in a weak dependence regime, which we formalize as follows. 
\ifcready
\else
A proof is provided in \appref{app:prop:svm:lr}.
\fi
\begin{proposition}
  \label{prop:svm:lr}
  Suppose $\rhomax = O(\eps)$ for some $\eps > 0$. For SVM and logistic regression applied on feature $f\colon \cX \to \mathbb{R}^d$ with covariance $\Lambda_{\ft} = I_d$, the optimal parameters satisfy
  \begin{align*}
    \wlr &= 2\lambda \cdot  \wsvm + o(\eps),\\
    \blr &= 2\lambda  \cdot \bsvm + o(\eps),
  \end{align*}
  where $\lambda$ is the hyperparameter in SVM. In addition, we have  $ \yh_{\SVM}(x; f) = \yh_{\LR}(x; f)$  for  $\eps$ sufficiently small.
\end{proposition}

\begin{remark} 
  Since H-score can also be directly maximized by implementing the maximal correlation regression \cite{xu2020maximal}, a similar connection holds for SVM and maximal correlation regression.
\end{remark}

\section{Fisher Kernel}  %

We demonstrate that
\emph{Fisher kernel} \cite{jaakkola1999exploiting, shawe2004kernel} can also be interpreted as a maximal correlation kernel.

Given a family of distributions $\pi(\cdot; \theta)$ supported on ${\cX}$ and parameterized by $\theta \in \mathbb{R}^m$,  suppose the score function $s_\theta(x) \defeq \frac{\partial }{\partial \theta} \log \pi(x; \theta)$ exists. Then, the Fisher kernel is defined as the projection kernel associated with the score function $s_\theta$, i.e., $\krf{s_\theta}$.

Specifically, we consider classification tasks where the joint distribution between data variable $X$ and label $Y$ are a mixture of the parameterized forms. Suppose for each class $Y = y \in \cY$, the data variable $X$ is generated from
\begin{align}
  P_{X|Y}(x|y) = \pi(x; \theta_y)
  \label{eq:assm:dist}
\end{align}
for some $\theta_y \in \mathbb{R}^m$. %
\ifcready
Then we have the following result.
\else
Then we have the following result, a proof of which is provided in \appref{app:thm:fisher:hgr}.
\fi
\begin{theorem}
  \label{thm:fisher:hgr}
  Suppose $\|\theta_y\| < \eps$ for all $y \in \cY$, and let $s(x)\defeq s_0(x)$. Then for the joint distribution $P_{X, Y} =  P_{X|Y}P_Y$ generated according to \eqref{eq:assm:dist}, we have
  \begin{gather}
    P_{X, Y}(x, y)
    = P_X(x)P_Y(y)\left(1 + \ip{s(x)}{\thetat_y}\right) + o(\eps),\label{eq:fisher:dcmp}\\
    \krf{s}(x, x') = \krs(x, x') + o(\eps),\label{eq:fisher:mck}
  \end{gather}
  where 
  $\tilde{\theta}_y \defeq \theta_y - \mathbb{E}[\theta_Y]$ denotes the centered $\theta_y$, and where  
  $\krs$ is the maximal correlation kernel defined on $P_{X, Y}$. In addition, the H-score of $s$ satisfies
  \begin{align}
    \Hs(s) 
    &= I(X; Y) + o(\eps^2),%
        \label{eq:i:xy}
  \end{align}
where $I(X; Y)$ denotes the mutual information between $X$ and $Y$.
\end{theorem}
From \eqref{eq:fisher:dcmp}, the score function $s$ is equal to the maximal correlation function $f^*$ of $P_{X, Y}$ up to a linear transformation [cf. \eqref{eq:modal}], and we have $P_{Y|X}(y|x) = \pkr{\krs}_{Y|X}(y|x) = \pkr{\krf{s}}_{Y|X}(y|x) + o(\eps)$. Therefore, Fisher kernel is the optimal kernel for tasks generated from \eqref{eq:assm:dist}.

\section{Conclusion}
\label{sec:conclusion}
In this paper, we study kernel methods from the perspective of feature subspace, where we demonstrate a connection between kernel methods and informative feature extraction problems. With SVM as an example, we illustrate the relationship between kernel methods and neural networks. The theoretical results can help guide practical kernel designs and incorporate kernel methods with feature-based learning approaches.

\section*{Acknowledgments}
This work was supported in part by the National Science Foundation (NSF) under Award CNS-2002908 and the Office of Naval Research (ONR) under grant N00014-19-1-2621.

\ifcready
\else

\clearpage

\appendix%
\subsection{Proof of \proptyref{propty:proj}}
\label{app:propty:proj}
  Suppose $\cG$ is a $d$-dimensional feature subspace. Let
 $\{g_1, \dots, g_d\}$ be
an orthonormal basis of $\cG$, i.e., we have $\ip{g_i}{g_j} = \kron_{\{i = j\}}$, and define $g(x) \defeq (g_1(x), g_2(x), \dots, g_d(x))^\T$. Then we have $\Lambda_{g} = I_d$ and $\krf{\cG}(x, x') = \ip{g(x)}{g(x')}$.

Therefore,
\begin{align*}
  [\tau(f)](x) 
  &= \E{\krf{\cG}(X, x)f(X)}\\
  &= \E{\ip{g(x)}{g(X)} \cdot f(X)}\\
  &= \sum_{i = 1}^d \E{ f(X) \cdot g_i(X)} \cdot g_i(x)\\
  &= \sum_{i = 1}^d \ip{f}{g_i}_{\spc{\cX}} \cdot g_i(x),
\end{align*}
which implies that $\tau(f) = \sum_{i = 1}^d \ip{f}{g_i}_{\spc{\cX}} \cdot g_i$.

  From the orthogonality principle, it suffices to prove that $\ip{f - \tau(f)}{\gh}_{\spc{\cX}} = 0$ for all $f \in \spc{\cX}$ and $\gh \in \cG$. To this end, suppose $\gh = \sum_{i = 1}^d c_i \cdot g_i$ for some $c_1, \dots, c_d \in \mathbb{R}$. Then, we have
  \begin{align*}
    \ip{\tau(f)}{\gh}_{\spc{\cX}} 
    &= \fip{\sum_{i = 1}^d \ip{f}{g_i}_{\spc{\cX}} \cdot g_i}{\sum_{j = 1}^d c_j \cdot g_j}_{\spc{\cX}} \\
    &= \sum_{i = 1}^d\sum_{j = 1}^d\ip{f}{g_i}_{\spc{\cX}} \cdot c_j \cdot \fip{  g_i}{g_j}_{\spc{\cX}} \\
    &= \sum_{i = 1}^d\sum_{j = 1}^d\ip{f}{g_i}_{\spc{\cX}} \cdot c_j \cdot \kron_{\{i = j\}}\\
    &= \sum_{i = 1}^d c_i \cdot \ip{f}{g_i}_{\spc{\cX}} \\
   &=  \fip{f}{\sum_{ i = 1}^d c_i \cdot g_i}_{\spc{\cX}} \\
   &=  \ip{f}{\gh}_{\spc{\cX}},
  \end{align*}
  which completes the proof.\hfill\QED

\subsection{Proof of \propref{prop:h:subspace}}
\label{app:prop:h:subspace}
  First, note that for each $y \in \cY$,
  \begin{align*}
    \E{\ft(X)\middle|Y = y}
    &= \E{f(X)\middle|Y = y} - \E{f(X)} \\
    &= \sum_{x \in \cX} [P_{X|Y}(x|y) - P_X(x)] \cdot f(x).
  \end{align*}
  Therefore, we have
  \begin{align*}
    &2 \cdot \Hs(f)\\
    &\quad=  \E{\bbnorm{\E{\Lambda_{f}^{-\frac12}\ft(X)\middle|Y}}^2}\\
    &\quad= \sum_{y \in \cY} P_Y(y) \cdot 
      \left(\E{\ft(X)\middle|Y = y}\right)^\T
      \La_{f}^{-1} \E{\ft(X)\middle|Y = y}\\
    &\quad= \sum_{y\in \cY} P_Y(y) \cdot \sum_{x \in \cX} \sum_{x' \in \cX} 
      \biggl([P_{X|Y}(x|y) - P_X(x)] \\
    &\quad\qquad \cdot
       \left(f^\T(x)  \La_f^{-1}  f(x') \right)\cdot
      [P_{X|Y}(x'|y) - P_X(x')]\biggr)\\
    &\quad= \Ed{P_{X X'}}{\krf{\cG}(X, X')} - \Ed{P_{X}P_{X'}}{\krf{\cG}(X, X')},
  \end{align*}
  which completes the proof.\hfill\QED

\subsection{Proof of \propref{prop:info:f}}
\label{app:prop:info:f}
  We start with the first equality. Suppose $P_{X, Y}$ has modal decomposition (cf. \propref{prop:modal}) 
  \begin{align*}
    P_{X, Y}(x, y) = P_X(x)P_Y(y)\cdot\left[1 + \rhomax \cdot f^*(x) \cdot g^*(y)\right], 
  \end{align*}
  where $g^*$ satisfies $\E{g^*(Y)} = 0$ and $\E{(g^*(Y))^2} = 1$.

  Then, we obtain
  \begin{align*}
    P_{X|Y=y}(x) = P_X(x)\cdot\left[1 + \rhomax \cdot f^*(x) \cdot g^*(y)\right],
  \end{align*}
  and the $P_{X, X'}$ as defined in \eqref{eq:p:xx} can be expressed as 
  \begin{align*}
    &P_{X X'}(x, x')\\
    &\qquad=\sum_{y \in \Y} P_{Y}(y)P_{X|Y=y}(x)P_{X|Y=y}(x')\\
    &\qquad=P_X(x)P_X(x') \cdot \Biggl[\sum_{y \in \Y} P_{Y}(y)\left(1 + \rhomax \cdot f^*(x) \cdot g^*(y)\right) \\
    &\qquad\qquad \cdot (1 + \rhomax \cdot f^*(x') \cdot g^*(y))\Biggl]\\
    &\qquad= P_X(x)P_X(x') \cdot \left[1 + \rhomax^2 \cdot f^*(x) \cdot f^*(x')\right],
  \end{align*}
  where to obtain the last equality follows from the fact that $\E{g^*(Y)} = 0, \E{(g^*(Y))^2} = 1$.

Note that since $P_{X'} = P_{X}$, we have
\begin{align}
  & P_{X X'}(x, x') - P_{X}(x)P_{X'}(x')\notag\\
  &\qquad=P_{X X'}(x, x') - P_{X}(x)P_{X}(x')\notag\\
  &\qquad= P_{X}(x)P_{X'}(x') \cdot \rhomax^2 \cdot f^*(x) \cdot f^*(x').
    \label{eq:p:xx:t}
\end{align}
In addition, let  $\tau \lra \krf{\cG}$ denote the  operator associated with $\krf{\cG}$, then from \proptyref{propty:proj} we have   $\tau(f^*) = \proj{f^*}{\cG}$. In addition, 
from the orthogonality principle, we have $\ip{f^* - \tau(f^*)}{\tau(f^*)}_{\spc{\cX}} = 0$ and thus
\begin{align}
  \ip{f^*}{\tau(f^*)}_{\spc{\cX}}
  &=   \ip{\tau(f^*)}{\tau(f^*)}_{\spc{\cX}} + \ip{f^* - \tau(f^*)}{\tau(f^*)}_{\spc{\cX}}\notag\\
  &= \ip{\tau(f^*)}{\tau(f^*)}_{\spc{\cX}}\notag\\
  &= \bnorm{\tau(f^*)}^2_{\spc{\cX}} = \bnorm{\proj{f^*}{\cG}}^2_{\spc{\cX}}.
    \label{eq:ip:tau}
\end{align}
Hence, the first equality of \eqref{eq:f:cG} can be obtained from
  \begin{align*}
    \Hs\left(\cG\right)
    &= \frac{\rhomax^2}2 \cdot \left(\Ed{P_{X X'}}{\krf{\cG}(X, X')} - \Ed{P_{X}P_{X'}}{\krf{\cG}(X, X')}\right)\\
    &=\frac{\rhomax^2}2 \cdot \Ed{P_{X}P_{X'}}{f^*(X) \cdot f^*(X') \cdot \krf{\cG}(X, X')}\\
    &=\frac{\rhomax^2}2 \cdot \sum_{x \in \cX} P_X(x) \cdot f^*(x) \cdot \E{f^*(X) \cdot \krf{\cG}(X, x)}\\
    &=\frac{\rhomax^2}2 \cdot \sum_{x \in \cX} P_X(x) \cdot f^*(x) \cdot [\tau(f^*)](x)\\
    &=\frac{\rhomax^2}2 \cdot \ip{f^*}{\tau(f^*)}_{\spc{\cX}}\\
    &=\frac{\rhomax^2}2 \cdot \bnorm{\proj{f^*}{ \cG}}_{\spc{\cX}}^2,
  \end{align*}
  where the first equality follows from \propref{prop:h:subspace},
  where the second equality follows from \eqref{eq:p:xx:t},
  and where the last equality follows from \eqref{eq:ip:tau}.

  To obtain the second and third equalities of \eqref{eq:f:cG}, it suffices to note that for all $f \in \cG$, we have
  \begin{align*}
    &\bnorm{\proj{f^*}{ \spn\{f\}}}_{\spc{\cX}}^2\\
    &\qquad = \bnorm{f^*}^2_{\spc{\cX}} - \bnorm{f^* - \proj{f^*}{ \spn\{f\}}}_{\spc{\cX}}^2\\
    &\qquad \leq \bnorm{f^*}^2_{\spc{\cX}} - \bnorm{f^* - \proj{f^*}{ \cG}}_{\spc{\cX}}^2\\
    &\qquad = \bnorm{\proj{f^*}{\cG}}_{\spc{\cX}}^2
  \end{align*}
  where the equalities follows from the orthogonality principle, and where the inequality follows from the definition of projection [cf. \eqref{eq:def:proj}]. In addition, it can be verified that the inequality holds with quality when $f = \proj{f^*}{\cG}$. 
  
  Hence,  for all $f \in \cG$, we have
  \begin{align*}
    \frac{\Hs(f)}{\Hs(\cG)} &= \frac{\Hs(\spn\{f\})}{\Hs(\cG)}\\
                            &= \frac{\bnorm{\proj{f^*}{ \spn\{f\}}}_{\spc{\cX}}^2}{\bnorm{\proj{f^*}{ \cG}}_{\spc{\cX}}^2}\leq 1 = \frac{\Hs(\proj{f^*}{\cG})}{\Hs(\cG)},
  \end{align*}  
  which completes the proof. 
\hfill\QED

\subsection{Proof of \proptyref{propty:kdm}}
\label{app:propty:kdm}
  It suffices to prove that $P_{Y|X} = P^{(\krs)}_{Y|X}$. To this end, suppose $P_{X, Y}$ satisfies the modal decomposition \eqref{eq:modal}, and let $f^*(x)\defeq (f_1^*(x), \dots, f_K^*(x))^\T, g^*(y) \defeq (g_1^*(y), \dots, g_K^*(y))^\T$, $\Sigma \defeq \diag(\sigma_1, \dots, \sigma_K)$. Then, it can be verified that  $\E{f_i^*(X)\middle|Y = y} = \sigma_i \cdot g_i^*(y)$ for all $i = 1, \dots, K$, which implies that  $\E{f^*(X)\middle|Y = y} = \Sigma \cdot g^*(y)$.

  Since $\La_{f^*} = I$, we have $\krs(x, x') = \ip{f^*(x)}{f^*(x')}$, for all $x, x' \in \cX$. 

  Hence, for all $x \in \cX, y \in \cY$, we have
  \begin{align*}
    P^{(\krs)}_{Y|X}(y|x)
    &= P_Y(y) \cdot \left( 1 + \E{\krs(X, x)\middle|Y = y}\right)\\
    &= P_Y(y) \cdot \left( 1 + \E{\ip{f^*(X)}{f^*(x)}\middle|Y = y}\right)\\
    &= P_Y(y) \cdot \left( 1 + \ip{\E{f^*(X)\middle|Y = y}}{f^*(x)}\right)\\
    &= P_Y(y) \cdot \left( 1 + \ip{\Sigma \cdot g^*(y)}{f^*(x)}\right)\\
    &= P_Y(y) \cdot \left( 1 + \sum_{i = 1}^K \sigma_i \cdot f_i^*(x)\cdot g_i^*(y)\right)\\
    &= P_{Y|X}(y|x),
  \end{align*} 
  which completes the proof.\hfill\QED

  \subsection{Proof of \proptyref{propty:map}}
\label{app:propty:map}
Our proof will make use of the following fact.
\begin{fact}
\label{fact:modal}
If $\cY = \{-1, 1\}$ and $P_Y(y)\equiv \frac12$, the modal decomposition of $P_{X, Y}$ (cf. \propref{prop:modal}) can be written as
\begin{align*}
  P_{X, Y}(x, y) = P_X(x)P_Y(y) \left(1 + \rhomax \cdot f^*(x) \cdot y \right),
\end{align*}
where $\rhomax$ is the maximal correlation coefficient, and $f^*$ is the maximal correlation function with $\E{(f^*(X))^2} = 1$.
\end{fact}

  From \factref{fact:modal}, we have
  \begin{align*}
    P_{Y|X}(y|x) 
    &= P_Y(y) \left(1 + y \cdot \rhomax \cdot f^*(x)\right)\\
    &= \frac12 \cdot \left(1 + y \cdot \rhomax \cdot f^*(x)\right),
  \end{align*}
  which implies that
  \begin{align*}
    \yh_\MAP(x)&= \argmax_{y \in \cY} P_{Y|X}(y|x)\\
    &= \argmax_{y \in \cY}~ \left[y \cdot f^*(x)\right]  = \sign(f^*(x)).
  \end{align*}
\hfill\QED

\subsection{Proof of \thmref{thm:svm}}
\label{app:thm:svm}
Our proof will make use of the following simple fact.
\begin{fact}
  \label{fact:bd:pos}
  Given a random variable $Z$ taking values from $\cZ$, let $z_{\min}$ denote the minimum entry in $\cZ$, then we have
\begin{align*}
  \E{Z} \leq \E{Z^+} \leq \E{Z} + (z_{\min})^{-},
\end{align*}
where $x^{-} \defeq \max\{-x, 0\}$.
\end{fact}

From \factref{fact:bd:pos}, we obtain
\begin{align}
  &\E{\ellh(Y, \ip{w}{f(X)} + b)}\notag\\
  &\quad= \E{\left( 1 - Y\cdot \fip{w}{f(X)} - Y\cdot b\right)^+}\notag\\
  &\quad\geq 1 - \fip{w}{\E{Y\cdot f(X)}} -  \E{Y} \cdot b\notag\\
  &\quad= 1 - \fip{w}{\E{Y\cdot f(X)}}.
\end{align}
Therefore, for all $w, b$, we have 
\begin{align*}
  \lsvm(f, w, b; \lambda)
  &= \E{\ellh(Y, \ip{w}{f(X)} + b)} + \frac{\lambda}{2}\cdot \norm{w}^2\\
  &\geq 1 - \fip{w}{\E{Y\cdot f(X)}} + \frac{\lambda}{2}\cdot \norm{w}^2\\
  &= 1 - \frac{1}{2\lambda} \cdot  \bbnorm{\E{f(X)Y}}^2\\
  &\qquad+ \frac{\lambda}{2} \cdot \bbnorm{w - \frac1\lambda\cdot\E{Y\cdot f(X)}}^2\\
  &\geq 1 - \frac{1}{2\lambda} \cdot  \bbnorm{\E{f(X)Y}}^2\\
  &= \lsvms(f; \lambda).
\end{align*}
Hence, we have
\begin{align*}
  \lsvm^*(f; \lambda) =   \lsvm(f; \wsvm, \bsvm, \lambda) \geq \lsvms(f; \lambda).
\end{align*}

Let $w' \defeq \frac1\lambda\cdot\E{Y\cdot f(X)}, b' \defeq - \ip{w'}{\E{f(X)}}$, then we have
\begin{align*}
\ip{w'}{f(X)} + b' =  \fip{w'}{\ft(X)}.
\end{align*}
Therefore, from the upper bound in \factref{fact:bd:pos}, we have
\begin{align}
  &\E{\ellh(Y, \ip{w'}{f(X)} + b')}\notag\\
  &\qquad\leq  1 - \E{Y \cdot \fip{w'}{\ft(X)}}\notag\\
  &\qquad\qquad+ \left(
  \min_{i} \left\{1 - y_i\cdot \fip{w'}{\ft(x_i)}\right\}
  \right)^{-}\notag\\
  &\qquad\leq  1 - \fip{w'}{\E{Y \cdot \ft(X)}}+ \left(1 - \frac{\lambda_\T}{\lambda}\right)^{-}\\
  &\qquad=  1 - \lambda \bbnorm{w'}^2+ \left(1 - \frac{\lambda_\T}{\lambda}\right)^{-}
\end{align}
where to obtain the last inequality, we have used the fact that
\begin{align*}
     \min_{i}\left\{1 - y_i \cdot \fip{w'}{\ft(x_i)}\right\} \geq 1 - \frac{\lambda_{\T}}{\lambda} 
\end{align*}
since for each $i$, we have
\begin{align*}
   1 - y_i \cdot \fip{w'}{\ft(x_i)}
  &\geq 1 - \bbabs{\fip{w'}{\ft(x_i)}}\\
  &\geq 1 - M \cdot \norm{w'}\\
  &= 1 - \frac{M}{\lambda} \cdot \bnorm{\E{Y\cdot f(X)}}\\
  &= 1 - \frac{\lambda_{\T}}{\lambda} .
\end{align*}
Therefore
\begin{align*}
  \lsvm^*(f; \lambda)
  &= \min_{w, b}\lsvm(f, w, b; \lambda)\\
  &\leq \lsvm(f, w', b'; \lambda)\\
  &\leq
    1 - \lambda \bbnorm{w'}^2+ \left(1 - \frac{\lambda_\T}{\lambda}\right)^{-} + \frac{\lambda}{2} \bbnorm{w'}^2\\
  &\leq
    1 - \frac{\lambda}{2} \bbnorm{w'}^2 + \left(1 - \frac{\lambda_\T}{\lambda}\right)^{-}\\
  &= \lsvms(f; \lambda)  + \left(1 - \frac{\lambda_\T}{\lambda}\right)^{-}\\
  &= \lsvms(f; \lambda)  + \left( \frac{\lambda_\T}{\lambda}- 1\right)^{+}.
\end{align*}

Finally, if $\lambda \geq \lambda_T$, it can be readily verified that $\lsvm^*(f; \lambda) = \lsvms(f; \lambda) = \lsvm(f, w', b'; \lambda)$. As a result, the optimal solution is given by
\begin{gather*}
  \wsvm = w' = \frac1\lambda\cdot\E{Y\cdot f(X)}, \\
  \bsvm = b'= - \ip{w'}{\E{f(X)}} = - \ip{w^*}{\E{f(X)}},
\end{gather*}
and we have
  \begin{align*}
    \fip{\wsvm}{f(x)} + \bsvm
    &= \ip{\wsvm}{\ft(x)}\\
    &= \frac1\lambda\cdot \fip{\E{Y\cdot f(X)}}{\ft(x)}\\
    &= \frac1\lambda\cdot \fip{\E{\ft(X) \cdot Y }}{\ft(x)}.
  \end{align*}
  Therefore, the SVM prediction is given by
  \begin{align*}
    \yh_\SVM(x; f, \lambda) &= \sign(\fip{\wsvm}{f(x)} + \bsvm)\\
                            &= \sign\left(\fip{\E{\ft(X)  Y }}{\ft(x)}\right).
  \end{align*}

  To obtain \eqref{eq:svm:decision:ncc}, note that for each $x \in \cX$, we have
  \begin{align*}
    &\norm{f(x) - \E{f(X)|Y = -1}}^2 -  \norm{f(x) - \E{f(X)|Y = 1}}^2\\
    &\qquad= \bbnorm{\ft(x) - \E{\ft(X)\middle|Y = -1}}^2\\
    &\qquad\qquad-  \bbnorm{\ft(x) - \E{\ft(X)\middle|Y = 1}}^2\\
    &\qquad= \bbnorm{\ft(x) - \E{\ft(X)\middle|Y = 1}}^2\\
    &\qquad\qquad-  \bbnorm{\ft(x) - \E{\ft(X)\middle|Y = -1}}^2\\
    &\qquad= \fip{\ft(x)}{\E{\ft(X)\middle|Y = 1} - \E{\ft(X)\middle|Y = -1}}\\
    &\qquad\qquad + \bbnorm{\E{\ft(X)\middle|Y = -1}}^2 -  \bbnorm{\E{\ft(X)\middle|Y = -1}}^2\\
    &\qquad= 2 \cdot \fip{\ft(x)}{\E{\ft(X) Y}},
  \end{align*}
  where we have used the facts that
  \begin{align*}
    \E{\ft(X)Y} 
    &= \E{Y \cdot \E{\ft(X)|Y}}\\
    &= \frac12 \left(\E{\ft(X)\middle|Y=1} - \E{\ft(X)\middle|Y=-1}\right),
  \end{align*}
  and
  \begin{align*}
    0 &= \E{\ft(X)}\\
      &= \E{ \E{\ft(X)|Y}}\\
      &= \frac12 \left(\E{\ft(X)\middle|Y=1} + \E{\ft(X)\middle|Y=-1}\right).
  \end{align*}
\hfill\QED

\subsection{Proof of \corolref{cor:svm}}
\label{app:cor:svm}
From \thmref{thm:svm}, when $\lambda \geq \lambda_\T$, we have
\begin{align*}
  \lsvm^*(f; \lambda) = \lsvms(f; \lambda) = 1 - \frac1{2\lambda} \cdot \bbnorm{\E{f(X)\cdot Y}}^2.
\end{align*}
Therefore,  it suffices to prove that
    \begin{align}
      r_{\min} \cdot \Hs(\ft) 
      \leq
      \frac12\cdot\bbnorm{\E{f(X)Y}}^2 \leq r_{\max}\cdot \Hs(\ft).
      \label{eq:cor:to:prove}
    \end{align}    
    To this end, note that we have
    \begin{align*}
      \bbnorm{\E{f(X)Y}}^2 
     =\bbnorm{\E{\ft(X)Y}}^2 
      &= \E{\bbnorm{Y\cdot\E{\ft(X)Y}}^2} \\      &=\E{\bbnorm{\E{\ft(X)\middle|Y}}^2 },
    \end{align*}
    where the last equality follows from the fact that for zero-mean $f$ and $Y$ uniformly distributed on $\{1, -1\}$, we have $\E{f(X)|Y = y} = y\cdot \E{f(X) \cdot Y}$ for $y \in \cY$.

    In addition, for all $v \in \spn\left\{\ft(x)\colon x \in \cX\right\}$, we have
    \begin{align*}
      r_{\min}  \leq \frac{\norm{v}^2}{\bbnorm{\Lambda_{\ft}^{-\frac12} v}^2} \leq 
      r_{\max}. 
    \end{align*}
    Therefore, we obtain
    \begin{align}
       {r_{\min}}\cdot
      \bbnorm{\Lambda_{\ft}^{-\frac12} \E{\ft(X)\middle|Y}}^2
      &\leq
        \bbnorm{\E{\ft(X)\middle|Y}}^2\notag\\
      &\leq r_{\max}\cdot
        \bbnorm{\Lambda_{\ft}^{-\frac12} \E{\ft(X)\middle|Y}}^2.
        \label{eq:exp:ft}
    \end{align}
    Taking the expectation of \eqref{eq:exp:ft} over $P_{Y}$ yields \eqref{eq:cor:to:prove}. \hfill\QED

\subsection{Proof of \propref{prop:svm:lr}}
\label{app:prop:svm:lr}
  First, note that logistic regression can be regarded as a special case of softmax regression with the correspondences $\wlr = w(1) - w(-1), \blr = b(1) - b(-1)$, where $w(y) \in \mathbb{R}^d$ and $b(y) \in \mathbb{R}$ are the weights and bias for softmax regression, respectively.

  In addition, from \cite[Theorem 2]{xu2022information},  the centered weight $\wt(y)\defeq w(y) - \E{w(Y)}$ and $b(y)$ are
  \begin{align*}
    \wt(y) &= \La_{\ft}^{-1} \E{\ft(X)\middle|Y = y} + o(\eps)\\
           &= \E{\ft(X)\middle|Y = y} + o(\eps),\\
    b(y) &= - \ip{\E{f(X)}}{\wt(y)} + o(\eps).
  \end{align*}

  Therefore, we obtain
  \begin{align*}
    \wlr &= w(1) - w(-1)\\
         &= \wt(1) - \wt(-1)\\
         &= \E{\ft(X)\middle|Y = 1} - \E{\ft(Y)\middle|Y = -1} + o(\eps)\\
         &= 2 \cdot \E{Y \cdot \ft(X)} + o(\eps)\\
         &= 2\lambda \cdot \wsvm + o(\eps)
  \end{align*}
  and
  \begin{align*}
    \blr &= b(1) - b(-1)\\
         &= -\ip{\E{f(X)}}{\wt(1) - \wt(-1)} + o(\eps)\\
         &= -\ip{\E{f(X)}}{\wlr} + o(\eps)\\
         &= -2\lambda \cdot \ip{\E{f(X)}}{\wsvm} + o(\eps)\\
         &= 2 \lambda \cdot  \bsvm + o(\eps),
  \end{align*}
  which implies that
  \begin{align*}
    \ip{\wlr}{ f(x)} + \blr = 2\lambda \cdot(\ip{\wsvm}{ f(x)} + \bsvm) + o(\eps).
  \end{align*}  

  From \eqref{eq:yh:svm} and \eqref{eq:yh:lr}, we have $\yh_{\SVM}(x; f) = \yh_{\LR}(x; f)$ for  $\eps$ sufficiently small.  
\hfill\QED

\subsection{Proof of \thmref{thm:fisher:hgr}}
\label{app:thm:fisher:hgr}
  To begin, note that we have
\begin{align*}
  P_{X|Y}(x|y) &= \pi(x; \theta_y)\\
               &=  \pi(x; 0) + \fip{
                 \left.\frac{\partial }{\partial \theta} \pi(x; \theta)\right|_{\theta = 0}}{\theta_y} + o(\eps)\\
               &= \pi(x; 0) (1 + \ip{s(x)}{\theta_y})  + o(\eps),
\end{align*}
which implies that %
\begin{align*}
  P_X(x)
  &= \sum_{y \in \Y} P_{X|Y}(x|y) P_Y(y)\\
  &= \sum_{y \in \Y} \pi(x; \theta_y) P_Y(y)\\
  &= \pi(x; 0) (1 + \ip{s(x)}{\E{\theta_Y}}) + o(\eps).
\end{align*}

Therefore, we obtain
\begin{align*}
  \frac{P_{X|Y}(x|y)}{P_X(x)}
  &= \frac{1 + \ip{s(x)}{\theta_y}}{1 + \ip{s(x)}{\E{\theta_Y}}}
    + o(\eps)\\
  &= \left(1 + \ip{s(x)}{\theta_y}\right)\cdot\left[1 - \ip{s(x)}{\E{\theta_Y}} + o(\eps)\right]\\
  &= 1 + \bip{s(x)}{\thetat_y}
  + o(\eps),
\end{align*}
where we have used the fact that $\norm{\E{\theta_Y}} = O(\eps)$ since $\norm{\theta_y} = O(\eps)$ for all $y \in \cY$. 

Hence, we have
\begin{align}
  P_{X, Y}(x, y)
  &= P_{X|Y}(x|y)P_Y(y)\notag\\
  &= P_X(x)P_Y(y)\left(1 + \ip{s(x)}{\thetat_y}\right) + o(\eps).%
    \label{eq:p:fisher}
\end{align}

Without loss of generality, we assume $\La_s$ is non-singular, since otherwise we can reparameterize $\theta$ to a  vector with dimension less than $m$. Compare \eqref{eq:p:fisher} with \eqref{eq:modal}, we have $K = m$. In addition, there exists $A \in \mathbb{R}^{m \times m}$ such that
\begin{align}
  s(x) = A \cdot f^*(x) + o(\eps), \quad \text{for all }x \in \cX, \label{eq:s:f}
\end{align}

Then, we can readily verify \eqref{eq:fisher:mck} from definition. Finally, \eqref{eq:i:xy} follows from \eqref{eq:s:f} and the fact that
\begin{align*}
  \Hs(f^*) = \frac{1}{2}\sum_{i = 1}^K \sigma_i^2 = I(X; Y) + o(\eps^2),
\end{align*}
where the second equality follows from the modal decomposition of mutual information (see, e.g., \cite[Lemma 16]{huang2019universal}).
\hfill\QED

\fi
% \clearpage

  %
  %
  %
      %
%

%
%
%
%
%
\bibliographystyle{IEEEtran}
\bibliography{ref}

\end{document}